\documentclass[a4paper]{article}

\usepackage{INTERSPEECH2018}
\usepackage{multirow}
\usepackage{mathtools}
\usepackage{verbatim}

\title{Completely Unsupervised Phoneme Recognition by Adversarially Learning Mapping Relationships from Audio Embeddings}
\name{Da-Rong Liu, Kuan-Yu Chen, Hung-Yi Lee, Lin-shan Lee}
\address{
  National Taiwan University}
\email{\{givebirthday, gary840212, tlkagkb93901106\}@gmail.com, lslee@gate.sinica.edu.tw}

\begin{document}

\maketitle

\begin{abstract}
Unsupervised discovery of acoustic tokens from audio corpora without annotation and learning vector representations for these tokens have been widely studied. Although these techniques have been shown successful in some applications such as query-by-example Spoken Term Detection~(STD), the lack of mapping relationships between these discovered tokens and real phonemes have limited the down-stream applications. This paper represents probably the first attempt towards the goal of completely unsupervised phoneme recognition, or mapping audio signals to phoneme sequences without phoneme-labeled audio data. The basic idea is to cluster the embedded acoustic tokens and learn the mapping between the cluster sequences and the unknown phoneme sequences with a Generative Adversarial Network~(GAN). An unsupervised phoneme recognition accuracy of 36\% was achieved in the preliminary experiments. 
\end{abstract}


\section{Introduction}
With the rapid development of deep learning, remarkable achievements in supervised speech recognition has been obtained~\cite{chorowski2015attention,chiu2017state}, but primarily relying on massive annotated data for model training, which is costly and labor requiring.
In contrast, in the era of big data, huge quantities of audio corpora are available almost everywhere, but there is almost no way to annotate them or use them in the supervised paradigm.
This is why unsupervised approaches of speech processing, including automatically discovering acoustic tokens and learning representations and linguistic structures from unlabeled audio corpora is attractive, desirable and critical~\cite{kamper2017segmental}.

In unsupervised discovery of acoustic tokens, the typical approach is to segment acoustically similar audio signal patterns followed by clustering the obtained repeated patterns in the corpora~\cite{pattern_overview, park2008unsupervised,pattern_NMF_ICASSP12,Wang12icassp}. 
These approaches were proved very useful in tasks such as query-by-example spoken term detection~(STD)~\cite{lyzinski2015evaluation, zhang2012resource, chung2018unsupervised, chung2017unsupervised}. 

On the other hand, in representation learning, efforts were made in trying to encode variable-length audio segments into vectors with fixed dimensionality~\cite{levin2013fixed}. 
These vector representations have been shown useful for many applications, such as speaker identification~\cite{dehak2009support} and audio emotion classification~\cite{schuller2009interspeech}, in addition to spoken term detection~(STD)~\cite{lee2013enhanced, chen2013hybrid, norouzian2012exploiting}, because such vector representations can be applied to standard classifiers to determine the speaker, the emotion label, or whether the input query is included.
For spoken document retrieval, by representing audio segments as vectors, search can be much more efficient compared to template matching over audio frames~\cite{levin2013fixed, levin2015segmental, kamper2016deep}.
Deep learning approaches have been popularly used for such purposes, with LSTM-based sequence-to-sequence auto-encoder being a good example~\cite{cho-al-emnlp14,sutskever2014sequence}. 
By minimizing the reconstruction error for the input audio sequences, the embeddings for the audio segments can be extracted from the bottleneck layer of the model. It has been shown that vectors obtained in this way carry the information of phonetic structures for the audio segments~\cite{chung2016audio, shen2017language}.  

All the above approaches were not able to learn the mapping relationships between the human-defined phonemes and the automatically discovered acoustic tokens.
The lack of this mapping relationships have seriously limited the applications obviously because the semantics and core information of the speech signal are carried by the phonemes. 
Although such automatically discovered acoustic tokens have been successfully transcribed into text with the help of some extra annotations~\cite{bansal2017towards}, to our knowledge, no work has been reported to try to transcribe the discovered acoustic tokens directly into phonemes in an completely unsupervised way.
Motivated by the very impressive work of neural machine translation~(NMT) without parallel data~\cite{conneau2018word, lample2018unsupervised}, with the concept of generative adversarial network (GANs)~\cite{goodfellow2014generative}, mapping from these automatically discovered acoustic tokens into phonemes seems possible because it is also a kind of translation except on the acoustic level. 

In this paper, we propose a completely unsupervised phoneme recognition framework, in which only unparalleled or unrelated speech utterances and text sentences are needed in model training. 
The audio signals are automatically segmented into acoustic tokens and encoded into representative vectors. 
These representative vectors are then clustered, and each speech utterance is represented as a cluster index sequence. 
A mapping relationship GAN is then developed, in which a generator transforms each cluster index sequence into a predicted phoneme sequence, and a discriminator is trained to distinguish the predicted phoneme sequence from the real phoneme sequences collected from text sentences.
A phoneme recognition accuracy of 36\% was achieved on TIMIT testing set for a model trained with TIMIT audio training set and unrelated text corpus.

\begin{figure*}[t]
  \centering
  \includegraphics[width=\linewidth]{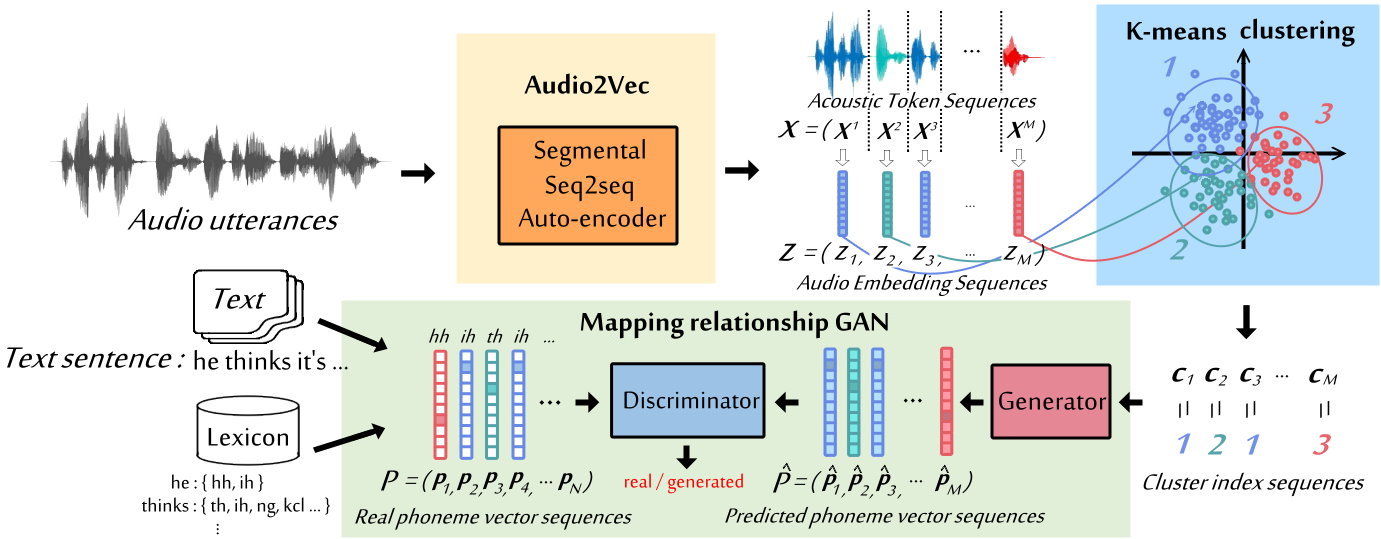}
  \caption{The proposed framework consists of three parts: Audio2Vec~(yellow) divides the audio utterances into segments and obtains the audio embeddings for the segments; all audio embeddings are then clustered by K-means~(blue) and assigned cluster indices; in mapping relationship GAN~(green), the generator produces predicted phoneme vector sequences from cluster index sequences, and the discriminator is trained to distinguish the predicted and the real phoneme vector sequences collected from text sentences and the lexicon.}
  \label{fig:model overview}
\end{figure*}

\section{Unsupervised phoneme recognition framework}

Figure \ref{fig:model overview} depicts the proposed framework including three parts: Audio2Vec, K-means clustering, and matching relationship GAN. 
Audio2Vec transforms each audio utterance into a sequence of audio embeddings.
Any unsupervised acoustic token discovery approach can be used here, not limited to the ones mentioned below.
All the audio embeddings are then K-means clustered and each assigned a cluster index. 
The mapping relationship GAN then learns to transform the cluster indices into phonemes without supervision.

\subsection{Audio2Vec}

Each utterance $\textbf{\uppercase{x}}$ is first divided into a sequence of automatically discovered acoustic tokens $\textbf{\uppercase{x}} = ( \uppercase{x}^{1}, \uppercase{x}^{2}, ..., \uppercase{x}^{M}  )$ in an unsupervised way, where $M$ is the number of tokens in the utterance. 
Each acoustic token $\uppercase{x}^{i}$ is then transformed into an audio embedding $z_{i} \in \mathbb{R}^{d}$, where $d$ is the dimensionality of the encoding space. 
So with Audio2Vec we encode the origin utterance into a vector sequence $\uppercase{z}=(z_{1},z_{2},...,z_{M})$, where $z_{i} \in \mathbb{R}^{d}$ is the audio embedding of the $i^{\text{~th}}$ token $\uppercase{x}^{i}$. 
Many available approaches can be used here for this purpose, but below we assume the recently developed Segmental Audio Word2Vec is used~\cite{wang2018segmental}. 
In this approach, automatic segmentation of utterances into acoustic tokens and audio embedding of the acoustic tokens into vector representations can be jointly learned in a Segmental Sequence-to-sequence Auto-encoder~(SSAE). In SSAE, a segmentation gate is inserted into the previously developed  Sequence-to-sequence Auto-encoder~(SAE).
The latter maps a variable-length audio segment into a fixed-length vector $ z \in \mathbb{R}^{d}$~\cite{shen2017language}.

SAE mentioned above consists of an RNN encoder and an RNN decoder.  
The RNN encoder reads the input sequence sequentially frame-by-frame and the hidden state of the RNN is updated accordingly. 
At the last frame of each acoustic token, the hidden state is taken as the audio embedding of the token and the hidden state is reset for the next token.
In addition to learning to segment the tokens in an utterance, the RNN encoder and decoder are trained by minimizing the mean squared reconstruction loss.
The whole training process requires no labeled data, but the audio embeddings are good representations for the tokens because based on them, the tokens can be reconstructed.

\subsection{K-means clustering}
In this stage, we collect all audio embeddings $z \in \mathbb{R}^{d}$ generated from all training audio data, and perform K-means to cluster them into $\uppercase{k}$ clusters. 
The choice of the parameter K will be discussed later on when discussing the experiments.
We further map the audio embedding sequence $\uppercase{z}=(z_{1},z_{2},...,z_{M})$ for each utterance to a cluster index sequence $\uppercase{c}=(c_{1},c_{2},...,c_{M})$, where $c_{i}=k\in [1,K]$ denotes the index for the cluster $z_{i}$ belongs to.
This clustering process reduces the difficult mapping between a high dimensional continuous embedding space and the phonemes to a simpler mapping between cluster indices and phonemes.

\subsection{Mapping relationship GAN} 

\begin{figure}[t]
  \centering
  \includegraphics[width=\linewidth]{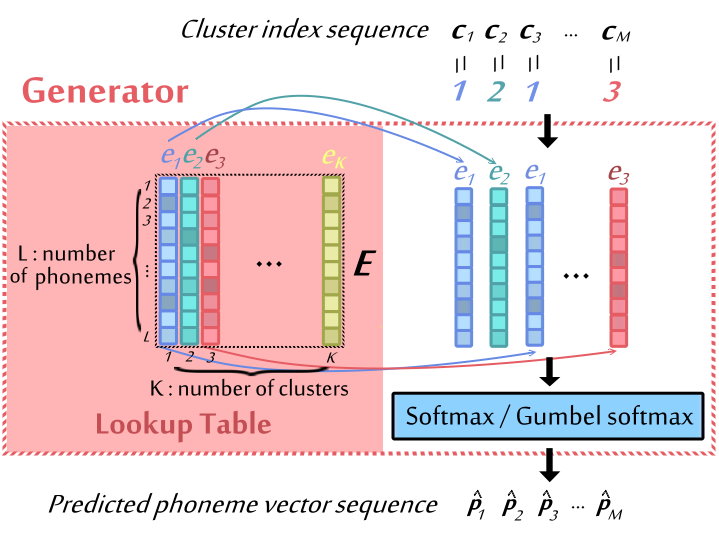}
  \caption{The generator in the mapping relationship GAN consists of a $K \times L$ matrix $E$, which is a lookup table, where K is the number of clusters and L the number of phonemes. Given a cluster index $c_{i}=k \in [1,K]$, we pick up the corresponding column $e_{k}$ in $E$ by table lookup, and perform softmax or gumbel-softmax to obtain the predicted phoneme vector.} 
  \label{fig:Generator}
\end{figure}

Generative Adversarial Networks~(GANs)~\cite{goodfellow2014generative, arjovsky2017wasserstein, yu2017seqgan}~is used in this stage. 
To learn the mapping relationship, we need a collection of text sentences, which does not have to be related to the audio.
Each of these text sentence is transformed into phoneme sequence based on a lexicon. 
Each phoneme is represented as a one-hot encoding vector $p$.
Hence a phoneme sequence for a sentence is represented as a vector sequence  $\uppercase{p}=(p_{1},p_{2},...,p_{N})$, where $N$ is the number of phonemes in the sentence.
$\uppercase{p}$ is referred to as 'real phoneme vector sequence' below. 

The mapping relationship GAN consist of a generator and a discriminator.
The generator G is an $K \times L$ matrix $E=\{e_{1}, e_{2},...,e_{K}\}$, where $L$ is the number of phonemes, $\uppercase{k}$ the number of clusters, as in Figure~\ref{fig:Generator}.
The $i^{\text{~th}}$ column of $E$, $e_{i} \in \mathbb{R}^{L}$, is intended to approximate the log probability of the $i^{\text{~th}}$ cluster of tokens over the $L$ phonemes, so $E$ is in fact a lookup table. 
Given a cluster index $c=k \in [1,K]$, the generator simply picks up the corresponding column of $E$, $e_{k}$, and performs a softmax function to produce a L-dimensional phoneme distribution $\hat{p}$.
In this way, for each cluster index sequence $C$ for an utterance, we can generate a sequence of vectors $\hat{\uppercase{p}}=G(\uppercase{c})=(\hat{p}_{1},\hat{p}_{2},...,\hat{p}_{M})$, each vector $\hat{p}_{i}$ for a token $\uppercase{x}^{i}$, referred to here as 'predicted phoneme vector sequence'. 
The discriminator D takes this predicted phoneme vector sequence $\hat{\uppercase{p}}$ and outputs a scalar.
The higher the scalar, the more possible it is a real phoneme vector sequence.

The discriminator is trained to distinguish between the predicted phoneme vector sequences $\hat{\uppercase{p}}$ and the real phoneme  vector sequences $\uppercase{p}$, while the generator is trained to cheat the discriminator. 
The discriminator is trained by minimizing the loss below following the concept of WGAN~\cite{arjovsky2017wasserstein},
\begin{equation}
  \mathcal{L}_{D} = -(\mathbb{E}_{\uppercase{p} \sim \mathbb{P}_{r}}[D(\uppercase{p})] - \mathbb{E}_{\hat{\uppercase{p}}=G(\uppercase{c}),  \uppercase{c}\sim \mathbb{P}_{C}}[D(\hat{\uppercase{p}})]), 
  \label{discriminator_loss} 
\end{equation}
where $\mathbb{P}_{r}$ is the distribution of the real phoneme vector sequences from text, $\mathbb{P}_{C}$ the distribution  of the cluster index sequences from audio data, and $\hat{\uppercase{p}}$ the output of generator, $\hat{\uppercase{p}}=G(\uppercase{c})$. 
Gradient penalty\cite{NIPS2017_7159} was applied to penalize functions having high gradient norm or changing too rapidly. 
 The generator loss is:
\begin{equation}
  \mathcal{L}_{G} = -\mathbb{E}_{\hat{\uppercase{p}}=G(\uppercase{c}),  \uppercase{c}\sim \mathbb{P}_{C}}[D(\hat{\uppercase{p}})].
  \label{generator_loss} 
\end{equation}
The generator and the discriminator are learned iteratively.
At the end, the generator is supposed to map any cluster index sequence to a phoneme vector sequence which 'looks like' a real one.
During testing, given an utterance, the cluster index sequence is generated and transformed into a phoneme vector sequence, and the phoneme with the highest score in each phoneme vector is taken as the phoneme recognition result.

In the generator, given a cluster index $c$, the predicted phoneme vector was obtained via softmax while the real phoneme vector is one-hot. An alternative way may be sampling from the phoneme distribution, but the sampling process is not differentiable. 
Therefore, we also try to use gumbel-softmax with straight-through estimator when training the generator~\cite{jang2016categorical, maddison2016concrete}.

\section{Experimental setup}
\subsection{Audio data}
The audio data were from TIMIT acoustic-phonetic corpus~\cite{garofolo1993darpa} including broadband recordings of phonetically-balanced read speech, with 6300 utterances from 630 speakers. 
The train/test sets were split with 462/168 non-overlapping speakers in each set respectively. 
Each utterance came with manually time-aligned phonetic and word transcriptions, as well as a 16-bit, 16kHz speech waveform file. 
The 39-dim MFCCs were extracted with utterance-wise cepstral mean and variance normalization (CMVN) applied. To slightly simplify our work, the start silence and end silence were removed from each utterance according to oracle phoneme boundaries.

\subsection{Lexicon and text data}
We built a lexicon containing all the words in the text of TIMIT training data. The text sentences were from the English monolingual data of WMT'16~\cite{bojar2016findings}, which was extracted from various online news published in 2015, containing roughly 27 millions of sentences. 
We only used thirty thousands sentences for which all words are in the above lexicon for training.

\subsection{Experimental setting}
All models were trained with stochastic gradient descent using a mini-batch size of 128, and Adam optimization technique with $\beta_{1}=0.9$, $\beta_{2}=0.999$ and $\epsilon=10^{-8}$. For Sequence-to-sequence Auto-Encoder, one-layer LSTM with 512 hidden units was used for both encoder and decoder. Training was set to 300 epoch with learning rate $5 \times 10^{-4}$. The cluster number for K-means ranged from 50 to 1000 and will be discussed below. The discriminator was a two-layer 1D CNN. The first layer concatenated 4 different kernel size: 3,5,7 and 9, each with 256 channels. The second layer was a convolution layer with kernel size 3 and 1024 channels. We used Leaky relu as the activation function.  
The learning rate of generator and discriminator were set to 0.01 and 0.001 respectively. Every training iteration consisted of 3 discriminator updates and a single generator update. The Gradient penalty ratio of the discriminator was set to 10~\cite{arjovsky2017wasserstein}. In  gumbel-softmax, the inverse temperature was set to 0.9 without annealing~\cite{maddison2014sampling}.

In the preliminary experiments, we didn't use SSAE mentioned above, but took the oracle phoneme boundaries provided in TIMIT for limit of time.
This is the only annotaion we used. The rest of the process were competely unsupervised. Automatically segmenting the utterances will be our future work. This gave about 160 thousands and 60 thousands segments in training/testing sets respectively. 
We trained all the models on the training set, and tested on both training and testing sets.
The evaluation metrics is phoneme accurary.

\section{Experimental results}
\begin{figure*}[t]
  \centering  \includegraphics[width=\linewidth]{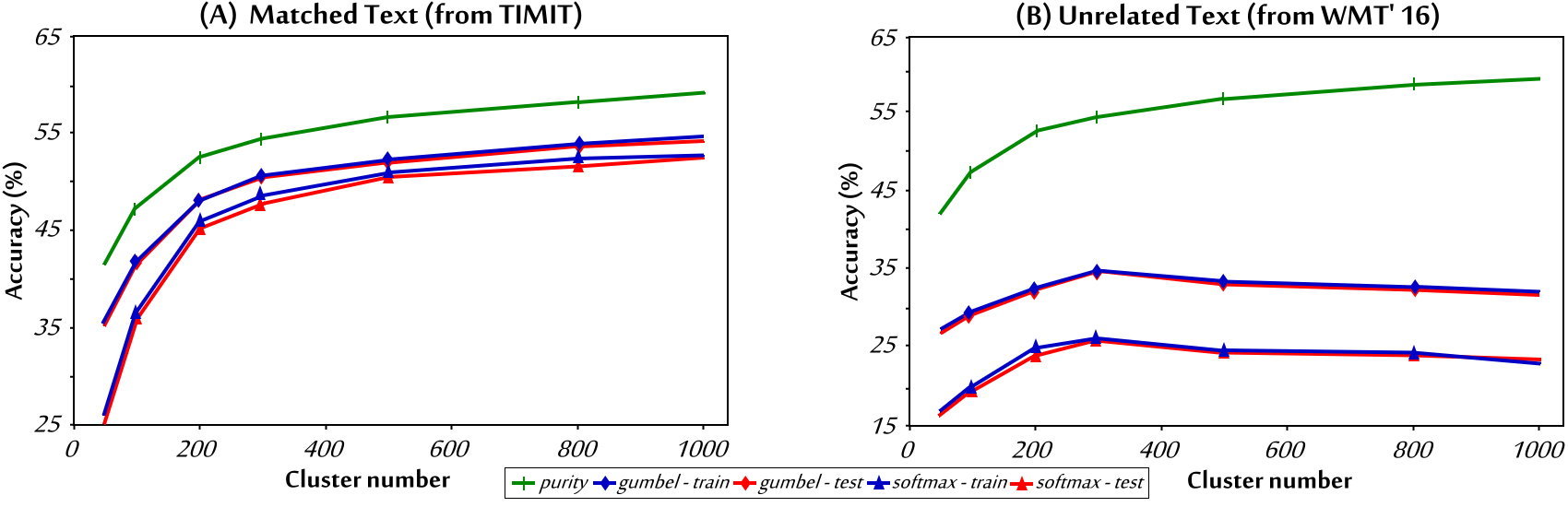}
  \caption{Phoneme accuracy for different numbers of clusters. Cluster purity~(green curves) are the upper bounds.}
  \label{fig:cluster_num_purity}
\end{figure*}

\subsection{Analysis on number of clusters}
We first varied the cluster number $\uppercase{k}$ to be 50, 100, 200, 300, 500, 800 and 1000. The results in Figure \ref{fig:cluster_num_purity} (A) are those for the reference phoneme sequences from TIMIT training set taken as the real phoneme sequences.
So the text sentences are the transcriptions of the audio utterances~(or extremely matched), not aligned though.
In Figure \ref{fig:cluster_num_purity} (B), the phoneme sequences generated from WMT'16 were used as real phoneme sequences, or unrelated to the audio utterances.
We obtained the correct phoneme of each segmented token based on the reference transcriptions, from which we defined the majority phoneme of each cluster.
The green curves in both Figure \ref{fig:cluster_num_purity} (A) and (B) are the cluster purity, which is the percentage of the majority phonemes in all clusters.  
This is the upper bound of the phoneme accuracy, if all segments in a cluster were mapped to its majority phoneme. Results for both softmax and gumbel softmax are shown for both training and testing sets.

In Figure \ref{fig:cluster_num_purity} (A), the performance increased with the cluster number. The best performance~(52.49\%) was achieved for 1000 clusters, very close to the upper bound or cluster purity of 59.02\%. 
Because the two domains were extremely matched, the mapping relationship GAN could identify the mapping relationship easily, so the performance was primarily determined  by the cluster purity, which increased with the cluster number.
In Figure \ref{fig:cluster_num_purity} (B), there was a much wider gap between the accuracy and the upper bound.
For 500 or more clusters (roughly 10 or more clusters for each phoneme in average), the mapping relationship became difficult to learn even though the cluster purity was higher for more clusters. 
The best performance happened around 300 clusters.

For all cases in Figure \ref{fig:cluster_num_purity} (A) and (B), gumbel-softmax outperformed softmax. 
The real phoneme sequences are all one-hot, but when using softmax, the generator naturally tried to make the predicted phoneme vectors close to one-hot to cheat the discriminator. But this is weird because a cluster mapped to a specific phoneme only when the cluster purity is 100\%. 
In contrast, with gumbel-softmax, the predicted phoneme vector was to be transformed towards one-hot. So the generator did not need to make the distribution very sharp, thus could concentrate more on making the predicted phoneme vector close to the real distributions.
Gumbel-softmax also made the training about 5 times faster.

\subsection{Unsupervised phoneme recognition accuracy}

The unsupervised phoneme recognition accuracy for 300 clusters in Figure \ref{fig:cluster_num_purity} is listed in Table \ref{table:comparison with baseline}.
Random baseline (row(a)) guessed uniformly from the 39 phoneme, while the most frequent phoneme baseline (row (b)) simply guessed the most frequent phoneme,/ix/,for all segments. 
Rows (c) to (f) are the results for the proposed approach for extremely matched data(Figure \ref{fig:cluster_num_purity} (A)) and unrelated data(Figure \ref{fig:cluster_num_purity} (B)) using softmax and gumbel-softmax
It is clear that even with unrelated text the proposed approach significantly outperformed the trivial baselines (rows (e), (f) v.s. (a), (b)).
For unrelated data, we also integrated the output of 6 models with slightly different parameters in the generator and discriminator by majority vote for both softmax and  gumbel-softmax  (rows (g), (h)). 
Obviously this improved the performance (rows (g) v.s. (e), (h) v.s. (f)), giving an accuracy exceeding 36\%. 
In all cases the training and testing results were very close. For unsupervised learning of phoneme characteristics, whether the model has seen the data is not important.

\begin{table}[t]
\centering
\caption{Comparison of different methods on training and testing data.}
\label{table:comparison with baseline}
\begin{tabular}{|l|l|l||c|c|}
\hline
\multicolumn{3}{|c||}{\multirow{2}{*}{\textbf{Methods}}} & \multicolumn{2}{c|}{\textbf{Accuracy}}     \\ \cline{4-5} 
\multicolumn{3}{|c||}{}& \multicolumn{1}{c|}{\textbf{Train}} &\multicolumn{1}{c|}{\textbf{Test}}  \\ \hline \hline
\multicolumn{3}{|c||}{(a) Random}                  & 2.56               & 2.56\\ \cline{1-5} 
\multicolumn{3}{|c||}{(b) Most Frequent Phoneme}   & 5.12   			  & 4.83\\ \cline{1-5} 
\multicolumn{1}{|c|}{\multirow{6}{*}{proposed}} & \multicolumn{1}{c|}{Matched}    & (c) Softmax & 48.58      & 48.23 \\ \cline{3-5} 
\multicolumn{1}{|c|}{}             				& \multicolumn{1}{c|}{(Figure \ref{fig:cluster_num_purity} (A))}  
& (d) Gumbel & 50.84  & 50.33 \\ \cline{2-5} 
\multicolumn{1}{|c|}{}             				& \multicolumn{1}{c|}{Unrelated}  & (e) Softmax & 25.78      & 25.68 \\ \cline{3-5} 
\multicolumn{1}{|c|}{}             				& \multicolumn{1}{c|}{(Figure \ref{fig:cluster_num_purity} (B))}  
& (f) Gumbel & 34.66 & 34.12 \\ \cline{2-5} 
\multicolumn{1}{|c|}{}             				& \multicolumn{1}{c|}{Unrelated}  & (g) Softmax & 31.35      & 31.21 \\ \cline{3-5} 
\multicolumn{1}{|c|}{}             				& \multicolumn{1}{c|}{(emsemble)} & (h) Gumbel  & \bf{36.27} & \bf{36.05} \\ \hline
\end{tabular}
\end{table}

\subsection{Comparison with supervised approaches} 
\begin{figure}[t]
  \centering
  \includegraphics[width=\linewidth]{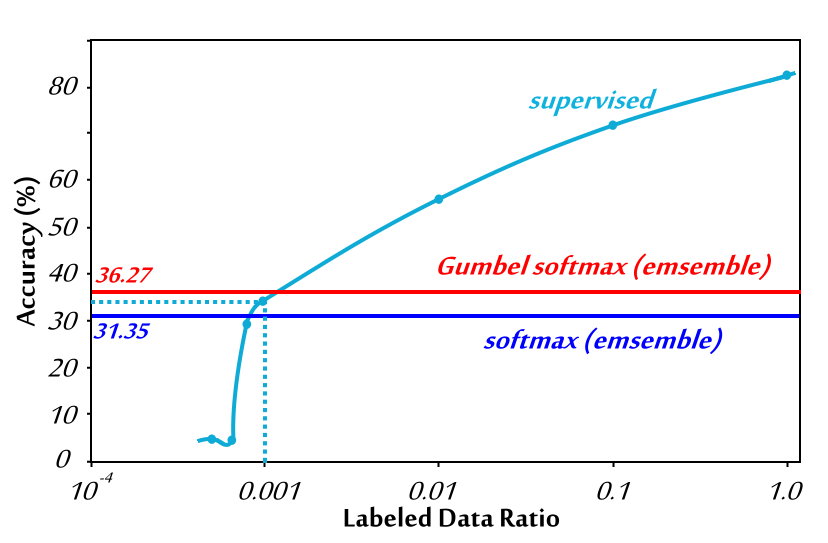}
  \caption{Comparison of unsupervised learning with unrelated text and supervised learning with different amounts of training data.}
  \label{fig:supervised}
\end{figure}

Here we wish to find out the accuracy we achieved can be obtained with how much of labeled data in supervised apprroach.
We used the labeled TIMIT training data to train a recurrent model to map each segment to its phoneme. 
We used a one-layer LSTM network with 512 hidden units with relu, cross-entropy as loss function.
The results are in Figure \ref{fig:supervised}, where the horizontal axis is the labeled data ratio, where 1.0 at the right end means we used exactly the same training data except all of them were labeled, and 0.01 means we only used 1\% of the data. 
Serious over-fitting occured at around 0.0007 of the ratio for the supervise model. 
We see the proposed unsupervised model with ensemble (the red line) exceeded the supervised model using 0.001~(0.1\%) of the training data but labeled. There is still a long way to go.

\section{Conclusions and future works}
In this work we proposed a framework to achieve compelety unsupervised phoneme recognition without parallel data. We demonstrate that the proposed unsupervised framework can achieve roughly comparable performance on supervised learning with 0.1\% of training data. There is still a long way to go along this direction, but here we show it is feasible to recognize phonemes without any labeled data.

\bibliographystyle{IEEEtran}
\bibliography{mybib,IR_bib}

\begin{thebibliography}{10}
\providecommand{\url}[1]{#1}
\csname url@samestyle\endcsname
\providecommand{\newblock}{\relax}
\providecommand{\bibinfo}[2]{#2}
\providecommand{\BIBentrySTDinterwordspacing}{\spaceskip=0pt\relax}
\providecommand{\BIBentryALTinterwordstretchfactor}{4}
\providecommand{\BIBentryALTinterwordspacing}{\spaceskip=\fontdimen2\font plus
\BIBentryALTinterwordstretchfactor\fontdimen3\font minus
  \fontdimen4\font\relax}
\providecommand{\BIBforeignlanguage}[2]{{%
\expandafter\ifx\csname l@#1\endcsname\relax
\typeout{** WARNING: IEEEtran.bst: No hyphenation pattern has been}%
\typeout{** loaded for the language `#1'. Using the pattern for}%
\typeout{** the default language instead.}%
\else
\language=\csname l@#1\endcsname
\fi
#2}}
\providecommand{\BIBdecl}{\relax}
\BIBdecl

\bibitem{chorowski2015attention}
J.~K. Chorowski, D.~Bahdanau, D.~Serdyuk, K.~Cho, and Y.~Bengio,
  ``Attention-based models for speech recognition,'' in \emph{Advances in
  neural information processing systems}, 2015, pp. 577--585.

\bibitem{chiu2017state}
C.-C. Chiu, T.~N. Sainath, Y.~Wu, R.~Prabhavalkar, P.~Nguyen, Z.~Chen,
  A.~Kannan, R.~J. Weiss, K.~Rao, K.~Gonina \emph{et~al.}, ``State-of-the-art
  speech recognition with sequence-to-sequence models,'' \emph{arXiv preprint
  arXiv:1712.01769}, 2017.

\bibitem{kamper2017segmental}
H.~Kamper, A.~Jansen, and S.~Goldwater, ``A segmental framework for
  fully-unsupervised large-vocabulary speech recognition,'' \emph{Computer
  Speech \& Language}, vol.~46, pp. 154--174, 2017.

\bibitem{pattern_overview}
J.~Glass, ``Towards unsupervised speech processing,'' in \emph{Information
  Science, Signal Processing and their Applications (ISSPA), 2012 11th
  International Conference on}, July 2012, pp. 1--4.

\bibitem{park2008unsupervised}
A.~S. Park and J.~R. Glass, ``Unsupervised pattern discovery in speech,''
  \emph{IEEE Transactions on Audio, Speech, and Language Processing}, vol.~16,
  no.~1, pp. 186--197, 2008.

\bibitem{pattern_NMF_ICASSP12}
J.~Driesen and H.~Van~hamme, ``Fast word acquisition in an {NMF}-based learning
  framework,'' in \emph{ICASSP}, 2012.

\bibitem{Wang12icassp}
H.~Wang, C.-C. Leung, T.~Lee, B.~Ma, and H.~Li, ``An acoustic segment modeling
  approach to query-by-example spoken term detection,'' in \emph{ICASSP}, 2012.

\bibitem{lyzinski2015evaluation}
V.~Lyzinski, G.~Sell, and A.~Jansen, ``An evaluation of graph clustering
  methods for unsupervised term discovery,'' in \emph{Sixteenth Annual
  Conference of the International Speech Communication Association}, 2015.

\bibitem{zhang2012resource}
Y.~Zhang, R.~Salakhutdinov, H.-A. Chang, and J.~Glass, ``Resource configurable
  spoken query detection using deep boltzmann machines,'' in \emph{Acoustics,
  Speech and Signal Processing (ICASSP), 2012 IEEE International Conference
  on}.\hskip 1em plus 0.5em minus 0.4em\relax IEEE, 2012, pp. 5161--5164.

\bibitem{chung2018unsupervised}
C.-T. Chung and L.-S. Lee, ``Unsupervised discovery of structured acoustic
  tokens with applications to spoken term detection,'' \emph{IEEE/ACM
  Transactions on Audio, Speech, and Language Processing}, vol.~26, no.~2, pp.
  394--405, 2018.

\bibitem{chung2017unsupervised}
C.-T. Chung, C.-Y. Tsai, C.-H. Liu, and L.-S. Lee, ``Unsupervised iterative
  deep learning of speech features and acoustic tokens with applications to
  spoken term detection,'' \emph{IEEE/ACM Transactions on Audio, Speech, and
  Language Processing}, vol.~25, no.~10, pp. 1914--1928, 2017.

\bibitem{levin2013fixed}
K.~Levin, K.~Henry, A.~Jansen, and K.~Livescu, ``Fixed-dimensional acoustic
  embeddings of variable-length segments in low-resource settings,'' in
  \emph{Automatic Speech Recognition and Understanding (ASRU), 2013 IEEE
  Workshop on}.\hskip 1em plus 0.5em minus 0.4em\relax IEEE, 2013, pp.
  410--415.

\bibitem{dehak2009support}
N.~Dehak, R.~Dehak, P.~Kenny, N.~Br{\"u}mmer, P.~Ouellet, and P.~Dumouchel,
  ``Support vector machines versus fast scoring in the low-dimensional total
  variability space for speaker verification,'' in \emph{Tenth Annual
  conference of the international speech communication association}, 2009.

\bibitem{schuller2009interspeech}
B.~Schuller, S.~Steidl, and A.~Batliner, ``The interspeech 2009 emotion
  challenge,'' in \emph{Tenth Annual Conference of the International Speech
  Communication Association}, 2009.

\bibitem{lee2013enhanced}
H.-y. Lee and L.-s. Lee, ``Enhanced spoken term detection using support vector
  machines and weighted pseudo examples,'' \emph{IEEE Transactions on Audio,
  Speech, and Language Processing}, vol.~21, no.~6, pp. 1272--1284, 2013.

\bibitem{chen2013hybrid}
I.-F. Chen and C.-H. Lee, ``A hybrid hmm/dnn approach to keyword spotting of
  short words.'' in \emph{INTERSPEECH}, 2013, pp. 1574--1578.

\bibitem{norouzian2012exploiting}
A.~Norouzian, A.~Jansen, R.~C. Rose, and S.~Thomas, ``Exploiting discriminative
  point process models for spoken term detection,'' in \emph{Thirteenth Annual
  Conference of the International Speech Communication Association}, 2012.

\bibitem{levin2015segmental}
K.~Levin, A.~Jansen, and B.~Van~Durme, ``Segmental acoustic indexing for zero
  resource keyword search,'' in \emph{Acoustics, Speech and Signal Processing
  (ICASSP), 2015 IEEE International Conference on}.\hskip 1em plus 0.5em minus
  0.4em\relax IEEE, 2015, pp. 5828--5832.

\bibitem{kamper2016deep}
H.~Kamper, W.~Wang, and K.~Livescu, ``Deep convolutional acoustic word
  embeddings using word-pair side information,'' in \emph{Acoustics, Speech and
  Signal Processing (ICASSP), 2016 IEEE International Conference on}.\hskip 1em
  plus 0.5em minus 0.4em\relax IEEE, 2016, pp. 4950--4954.

\bibitem{cho-al-emnlp14}
\BIBentryALTinterwordspacing
K.~Cho, B.~van Merri{\"{e}}nboer, {\c C}.~G{\"{u}}l{\c c}ehre, D.~Bahdanau,
  F.~Bougares, H.~Schwenk, and Y.~Bengio, ``Learning phrase representations
  using rnn encoder--decoder for statistical machine translation,'' in
  \emph{Proceedings of the 2014 Conference on Empirical Methods in Natural
  Language Processing (EMNLP)}.\hskip 1em plus 0.5em minus 0.4em\relax Doha,
  Qatar: Association for Computational Linguistics, Oct. 2014, pp. 1724--1734.
  [Online]. Available: \url{http://www.aclweb.org/anthology/D14-1179}
\BIBentrySTDinterwordspacing

\bibitem{sutskever2014sequence}
I.~Sutskever, O.~Vinyals, and Q.~V. Le, ``Sequence to sequence learning with
  neural networks,'' in \emph{Advances in neural information processing
  systems}, 2014, pp. 3104--3112.

\bibitem{chung2016audio}
Y.-A. Chung, C.-C. Wu, C.-H. Shen, H.-Y. Lee, and L.-S. Lee, ``Audio word2vec:
  Unsupervised learning of audio segment representations using
  sequence-to-sequence autoencoder,'' \emph{arXiv preprint arXiv:1603.00982},
  2016.

\bibitem{shen2017language}
C.-H. Shen, J.~Y. Sung, and H.-Y. Lee, ``Language transfer of audio word2vec:
  Learning audio segment representations without target language data,''
  \emph{arXiv preprint arXiv:1707.06519}, 2017.

\bibitem{bansal2017towards}
S.~Bansal, H.~Kamper, A.~Lopez, and S.~Goldwater, ``Towards speech-to-text
  translation without speech recognition,'' \emph{arXiv preprint
  arXiv:1702.03856}, 2017.

\bibitem{conneau2018word}
A.~Conneau, G.~Lample, M.~Ranzato, L.~Denoyer, and H.~J{\'e}gou, ``Word
  translation without parallel data,'' in \emph{Proceedings of the
  International Conference on Learning Representations (ICLR)}, 2018.

\bibitem{lample2018unsupervised}
G.~Lample, L.~Denoyer, and M.~Ranzato, ``Unsupervised machine translation using
  monolingual corpora only,'' in \emph{Proceedings of the International
  Conference on Learning Representations (ICLR)}, 2018.

\bibitem{goodfellow2014generative}
I.~Goodfellow, J.~Pouget-Abadie, M.~Mirza, B.~Xu, D.~Warde-Farley, S.~Ozair,
  A.~Courville, and Y.~Bengio, ``Generative adversarial nets,'' in
  \emph{Advances in neural information processing systems}, 2014, pp.
  2672--2680.

\bibitem{wang2018segmental}
Y.-H. Wang, H.-y. Lee, and L.-s. Lee, ``Segmental audio word2vec: Representing
  utterances as sequences of vectors with applications in spoken term
  detection,'' in \emph{Acoustics, Speech and Signal Processing (ICASSP), 2018
  IEEE International Conference}.\hskip 1em plus 0.5em minus 0.4em\relax IEEE,
  2018.

\bibitem{arjovsky2017wasserstein}
M.~Arjovsky, S.~Chintala, and L.~Bottou, ``Wasserstein gan,'' \emph{arXiv
  preprint arXiv:1701.07875}, 2017.

\bibitem{yu2017seqgan}
L.~Yu, W.~Zhang, J.~Wang, and Y.~Yu, ``Seqgan: Sequence generative adversarial
  nets with policy gradient.'' 2017.

\bibitem{NIPS2017_7159}
\BIBentryALTinterwordspacing
I.~Gulrajani, F.~Ahmed, M.~Arjovsky, V.~Dumoulin, and A.~C. Courville,
  ``Improved training of wasserstein gans,'' in \emph{Advances in Neural
  Information Processing Systems 30}, I.~Guyon, U.~V. Luxburg, S.~Bengio,
  H.~Wallach, R.~Fergus, S.~Vishwanathan, and R.~Garnett, Eds.\hskip 1em plus
  0.5em minus 0.4em\relax Curran Associates, Inc., 2017, pp. 5767--5777.
  [Online]. Available:
  \url{http://papers.nips.cc/paper/7159-improved-training-of-wasserstein-gans.pdf}
\BIBentrySTDinterwordspacing

\bibitem{jang2016categorical}
E.~Jang, S.~Gu, and B.~Poole, ``Categorical reparameterization with
  gumbel-softmax,'' \emph{arXiv preprint arXiv:1611.01144}, 2016.

\bibitem{maddison2016concrete}
C.~J. Maddison, A.~Mnih, and Y.~W. Teh, ``The concrete distribution: A
  continuous relaxation of discrete random variables,'' \emph{arXiv preprint
  arXiv:1611.00712}, 2016.

\bibitem{garofolo1993darpa}
J.~S. Garofolo, L.~F. Lamel, W.~M. Fisher, J.~G. Fiscus, and D.~S. Pallett,
  ``Darpa timit acoustic-phonetic continous speech corpus cd-rom. nist speech
  disc 1-1.1,'' \emph{NASA STI/Recon technical report n}, vol.~93, 1993.

\bibitem{bojar2016findings}
O.~Bojar, R.~Chatterjee, C.~Federmann, Y.~Graham, B.~Haddow, M.~Huck, A.~J.
  Yepes, P.~Koehn, V.~Logacheva, C.~Monz \emph{et~al.}, ``Findings of the 2016
  conference on machine translation.'' in \emph{ACL 2016 FIRST CONFERENCE ON
  MACHINE TRANSLATION (WMT16)}.\hskip 1em plus 0.5em minus 0.4em\relax The
  Association for Computational Linguistics, 2016, pp. 131--198.

\bibitem{maddison2014sampling}
C.~J. Maddison, D.~Tarlow, and T.~Minka, ``A* sampling,'' in \emph{Advances in
  Neural Information Processing Systems}, 2014, pp. 3086--3094.

\end{thebibliography}
\end{document}